\documentclass{edm_article}
\usepackage{multirow} 
\usepackage[protrusion]{microtype} 
\usepackage{csquotes} 
\usepackage{enumitem} 
\setitemize{itemsep=0pt,topsep=0pt}
\setenumerate{itemsep=0pt,topsep=0pt}
\usepackage[breaklinks,colorlinks,linkcolor=blue,urlcolor=blue,citecolor=blue]{hyperref} 

\begin{document}
\title{Ranking-Based At-Risk Student Prediction Using Federated Learning and Differential Features}

\numberofauthors{5} 
 
 \author{
 \alignauthor
 Shunsuke Yoneda\\
        \affaddr{Kyushu University}\\
        \email{yoneda.shunsuke.860@\\s.kyushu-u.ac.jp}
 \alignauthor
 Valdemar Švábenský \\
        \affaddr{Kyushu University}\\
        \email{valdemar@kyudai.jp}
 \alignauthor
 Gen Li\\
        \affaddr{Kyushu University}\\
        \email{gen.li@limu.ait.kyushu-u.ac.jp}
 \and
 Daisuke Deguchi\\
        \affaddr{Nagoya University}\\
        \email{ddeguchi@nagoya-u.ac.jp}
 \alignauthor
 Atsushi Shimada\\
        \affaddr{Kyushu University}\\
        \email{atsushi@limu.ait.kyushu-u.ac.jp}
 }

\toappear{\scriptsize S. Yoneda, V. Švábenský, G. Li, D. Deguchi, and A. Shimada. Ranking-Based At-Risk Student Prediction Using Federated Learning and Differential Features. In \textit{Proceedings of the 18th International Conference on Educational Data Mining}, July 2025. International Educational Data Mining Society.\\

© 2025 Copyright is held by the author(s). This work is distributed under the Creative Commons Attribution NonCommercial NoDerivatives 4.0 International (CC BY-NC-ND 4.0) license.\\
\url{https://doi.org/10.5281/zenodo.15870193}
}

\hyphenation{ana-lysis}
\hyphenation{de-monstrated}

\maketitle

\begin{abstract}
Digital textbooks are widely used in various educational contexts, such as university courses and online lectures. Such textbooks yield learning log data that have been used in numerous educational data mining (EDM) studies for student behavior analysis and performance prediction. However, these studies have faced challenges in integrating confidential data, such as academic records and learning logs, across schools due to privacy concerns. Consequently, analyses are often conducted with data limited to a single school, which makes developing high-performing and generalizable models difficult.
This study proposes a method that combines federated learning and differential features to address these issues. Federated learning enables model training without centralizing data, thereby preserving student privacy. Differential features, which utilize relative values instead of absolute values, enhance model performance and generalizability.
To evaluate the proposed method, a model for predicting at-risk students was trained using data from 1,136 students across 12 courses conducted over 4 years, and validated on hold-out test data from 5 other courses. Experimental results demonstrated that the proposed method addresses privacy concerns while achieving performance comparable to that of models trained via centralized learning in terms of Top-n precision, nDCG, and PR-AUC. Furthermore, using differential features improved prediction performance across all evaluation datasets compared to non-differential approaches. The trained models were also applicable for early prediction, achieving high performance in detecting at-risk students in earlier stages of the semester within the validation datasets.
\end{abstract}


\section{Introduction}
Digital textbooks are widely used due to their capability to not only allow students to access learning materials on personal devices but also collect records of their interactions as learning logs. These digital textbooks are now implemented in many educational institutions \cite{digital_textbooks1, digital_textbooks2, cadd_broad1, cadd_broad2}, leading to the accumulation of vast amounts of learning logs.

This development has motivated research on students' learning behavior and the prediction of academic performance using learning logs \cite{learning_log_data1, learning_log_data2, learning_log_data3}. For example, studies have developed systems that provide instructors with real-time visualizations of students' learning progress, such as the percentage of students keeping pace with the lecture or those remaining on a previous page \cite{learning_log_data}. Other studies have utilized learning log data to predict final exam scores and classify students into higher- and lower-performing \cite{grade2}.

However, these studies have faced major difficulties in integrating academic performance data and learning logs across schools, which introduces privacy concerns \cite{EDM_about_integrating,privacy_baker,privacy_mark}. Consequently, creating generalizable and high-performance models is challenging due to the limited availability of data \cite{relationship_dataset1,relationship_dataset2}. Thus, developing methods for performance prediction without directly integrating data is crucial to advance EDM research and strengthen learning support in educational practice.

In conventional machine learning (ML), data stored in separate locations must be centralized on a single server for model training, which raises concerns about privacy -- an important topic within the EDM community \cite{EDM_about_integrating,privacy_baker,privacy_mark}. Prior work has focused on this issue from the perspective of privacy-preserving EDM infrastructures, such as MORF \cite{MORF}, which allow to train ML models without direct access to the data.
\textit{Federated learning}, which has gained attention as a privacy preserving ML approach in non-EDM contexts \cite{applications_of_federated_learning,Owkin,financial,federated2}, supports privacy from a different perspective. It is based on distributing model parameters to data owners (hereinafter, referred to as \enquote{clients}), who train the model locally on their data. The locally trained parameters are then aggregated on a central server, enabling training without transferring raw data. 
Compared to approaches like MORF, the advantage of federated learning is that it supports privacy without the need to establish a centrally managed infrastructure. Since data can be processed in a decentralized manner, no management cost applies, and security measures associated with centralized data management are also eliminated.

However, when applying federated learning, discrepancies in the feature distributions among clients can arise due to differences in context, such as e-book usage frequencies or course schedules. These differences can bias model training and degrade the model performance \cite{Federated_learning_problem1, Federated_learning_problem2,Federated_learning_problem3}.
Prior studies have explored methods focused on improving aggregation to prevent these differences from degrading the models performance \cite{Federated_learning_modify1,Federated_learning_modify2}. Our study proposes an approach that addresses these discrepancies through data preprocessing using \textit{differential features} \cite{pairwise_difference1,pairwise_difference2}.
By utilizing relative feature values instead of absolute ones, this approach mitigates disparities in feature distributions among clients. As a result, it improves both generalizability and model performance -- two key aspects of many EDM studies \cite{generalizability_1,generalizability_2,generalizability_3,generalizability_4}.

In summary, federated learning with differential features has shown promise for preserving privacy and enhancing ML model properties in non-EDM contexts. However, to the best of our knowledge, this approach has not been validated in the context of EDM research. Therefore, our study aims to \textit{evaluate ML models for at-risk student prediction using the unique approaches of federated learning with differential features}. Furthermore, we investigate whether accurate predictions can be achieved in the early stages of a course—specifically, using learning log data collected up to the halfway point of the lecture sessions—as timely identification of at-risk students is essential for practical interventions in real-world educational settings. 

Our study makes the following contributions:
\begin{itemize} 
    \item We enable accurate prediction of students' academic performance using learning log data, with a particular focus on identifying \textit{at-risk students}, while preserving privacy through \textit{federated learning}.
    \item We enhance model generalizability and performance by utilizing \textit{differential features}, which capture relative differences in students' learning behaviors and academic performance instead of absolute values, thereby mitigating distributional disparities across datasets.
    \item We demonstrate the applicability of our approach to \textit{early prediction}, showing that at-risk students can be accurately identified in the early stages of a course based on partial learning log data.
\end{itemize}

\section{Related Work}
\label{related_work}
We aim to develop a generalized, high-performance ML model for predicting grades while preserving privacy using federated learning with differential features. Therefore, this section reviews prior studies related to this research across two themes: \textit{grade prediction} and \textit{federated learning}.

\subsection{Prediction of Students' Grades}
Predicting grades is a crucial area of EDM research aimed at supporting learning activities and enabling personalized educational interventions. Various methods have been proposed, including classification models that categorize students based on their predicted performance \cite{grade_prediction_classification1,grade_prediction_classification2,fEDM21student} and regression models that estimate continuous grade values \cite{grade_prediction_regression1,grade_prediction_regression2}.

For example, Chen et al. \cite{grade2} constructed a model to classify university students into \enquote{higher-score students} or \enquote{lower-score students} by leveraging various features, including learning behaviors within digital textbooks (e.g., turning pages, adding/deleting markers, and editing/removing memos).
Ong et al. \cite{instructor_related} examined whether incorporating \enquote{instructor-related features} improves the performance of students' grade prediction using both regression and classification.
Altabrawee et al. \cite{grade_prediction_computer_science} developed a model to predict academic performance in computer science courses at a university. Their model used features such as frequency of using the internet for studying, time spent on social media, and previous semester grades to classify students as either \enquote{Good} or \enquote{Weak}.

Overall, research on grade prediction encompasses a wide range of approaches, with each study employing different features and methods. However, many studies have focused on predicting grades for specific courses, which limits their applicability. The impact of data changes throughout a single course~\cite{sukrit_early_prediction} as well as across various courses. Yet, the models' generalizability to different course contents has not been sufficiently investigated. 
Therefore, this study aims to develop and evaluate a model capable of generalizing across courses to detect students with poor final grades. To achieve this, federated learning was utilized to address privacy concerns, and differential features were introduced to improve the model's generalizability and performance.

\subsection{Federated Learning}
Federated learning has gained considerable attention in recent years for enabling privacy-preserving model training across various fields outside EDM \cite{applications_of_federated_learning}.
Oldenhof et al. \cite{Owkin} demonstrating federated learning in the medical drug discovery process. Federated learning was employed to train predictive models collaboratively across multiple pharmaceutical companies without centralizing sensitive data. This approach yielded a shared predictive model while maintaining data confidentiality.
Kanamori et al.~\cite{financial} applied federated learning in detecting fraudulent financial transactions. They developed a model in collaboration with five banks, enabling data analysis without centralizing confidential information. The federated learning system outperformed models trained solely on individual bank data, achieving higher performance in detecting fraudulent transactions. This system enabled the detection of fraudulent accounts before actual monetary losses occurred.
Liu et al. \cite{federated2} proposed a federated learning approach for traffic flow prediction, allowing multiple organizations to collaboratively build predictive models without sharing raw data. By leveraging information collected locally, organizations could develop more accurate traffic prediction models while preserving data privacy. Each organization used its own traffic data to train models and shared only the resulting parameters, ensuring the protection of confidential information. Experiments using real-world traffic data demonstrated that this method achieved superior predictive performance compared to traditional approaches.

As demonstrated in the aforementioned studies, federated learning shows strong promise in other fields, enabling multiple organizations to collaboratively train ML models while preserving data privacy. Moreover, this approach increases the amount of data available for training, thereby potentially also enhancing predictive performance. 
However, federated learning remains underexplored in educational contexts. No publication at the EDM conference in the past five years (2020--2024) has focused on federated learning (based on examining the titles in the proceedings). Since the effectiveness of federated learning has not been thoroughly validated -- particularly in addressing challenges posed by heterogeneous data distributions -- our study is unique because it focuses on these aspects.

To the best of our knowledge, only one recent paper has employed federated learning in education. In 2024, Haastrecht et al. \cite{Federated_grade_prediction} investigated federated learning for educational analytics by comparing different approaches across multiple prediction tasks. While their study demonstrated the feasibility of federated learning, it did not address the impact of feature distribution discrepancies across clients, which can substantially affect model performance. 
By mitigating the issues caused by feature distribution discrepancies using differential features, our study explores the potential of federated learning in EDM and validates its effectiveness.

\section{The Proposed Modeling Method}
\begin{figure}[htbp]
\begin{center}
\Description{The image of proposed method} 
\includegraphics[width=\linewidth]{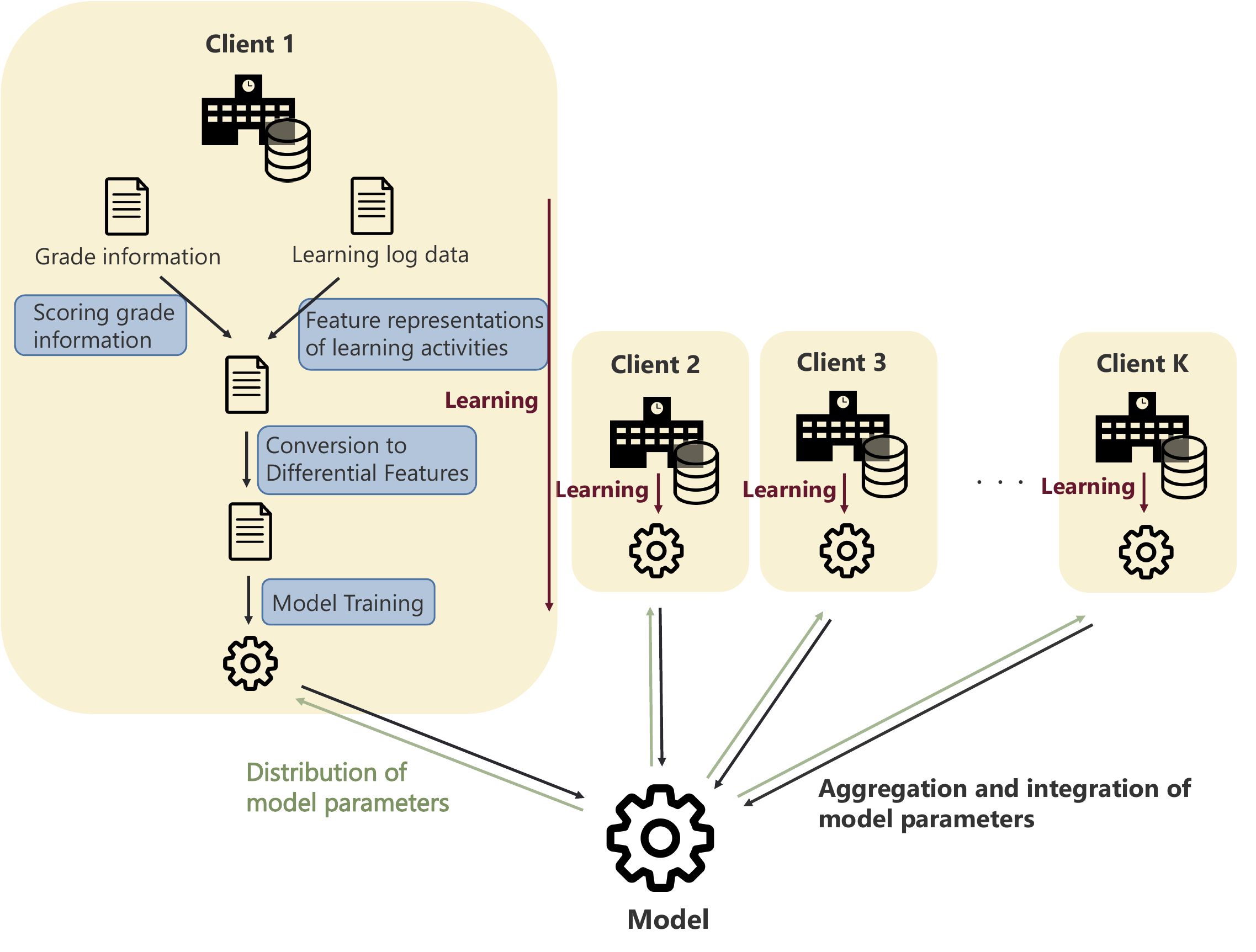}
\caption{The overall framework of our proposed method}
\label{fig:big_picture}
\end{center}
\end{figure}

This section explains our method for creating a model that preserves privacy through federated learning and leverages differential features. The overall framework is illustrated in Figure \ref{fig:big_picture}. The following subsections focus on \textit{federated learning} and \textit{differential features}, detailing how these approaches are used to conduct learning and prediction.

\subsection{Federated Learning}
\begin{figure}[htbp]
\begin{center}
\Description{The image of sequence diagram of learning and prediction in federated learning} 
\includegraphics[width=\linewidth]{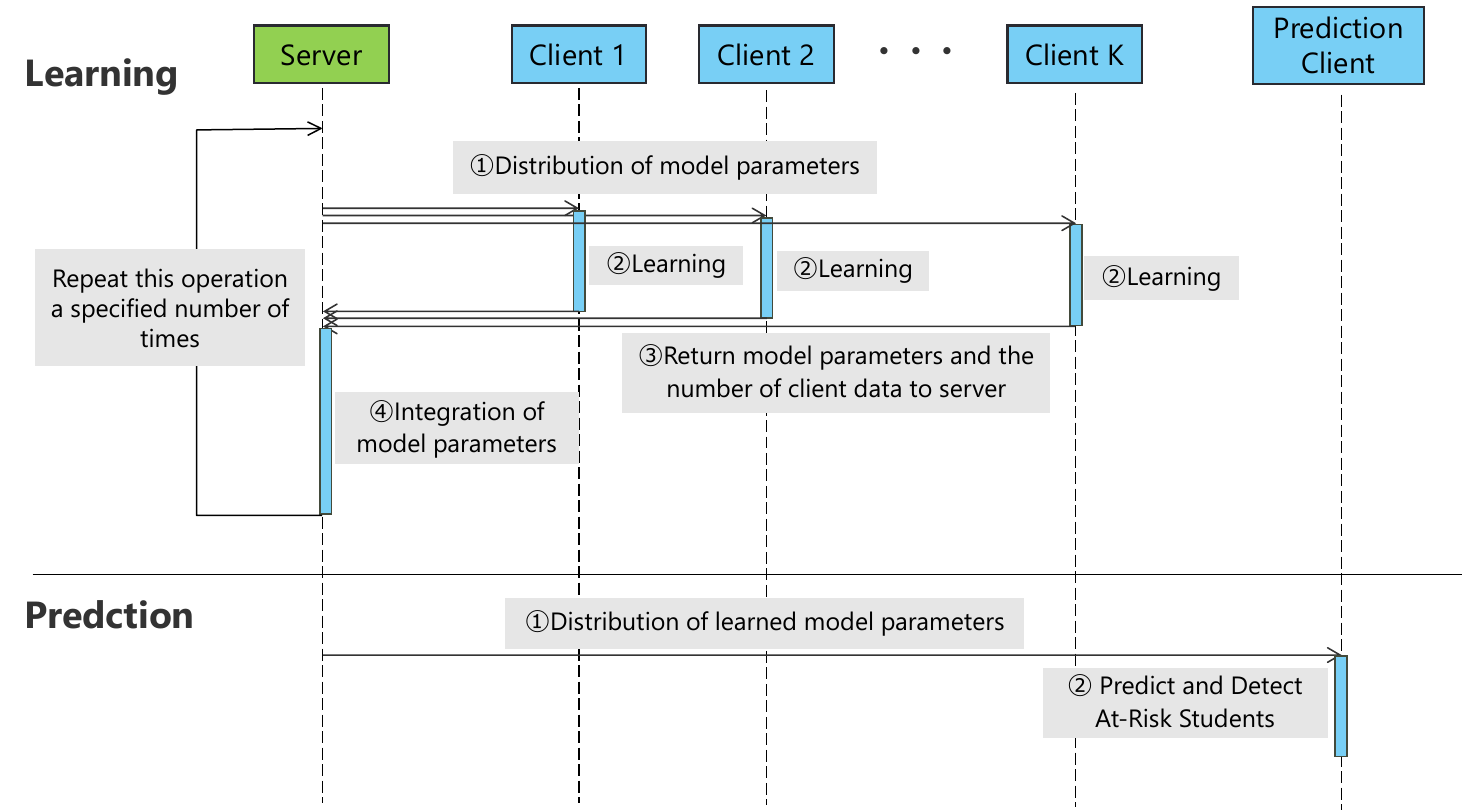}
\caption{The sequence diagram of learning and prediction in federated learning}
\label{fig:federated_learning_sequence}
\end{center}
\end{figure}

Figure \ref{fig:federated_learning_sequence} illustrates the learning and prediction processes of federated learning utilized in this study. This approach enables model training and prediction without the need to aggregate data on server, ensuring data privacy.

In the learning phase, server distributes model parameters to each client. Each client trains the model using its locally held data. Subsequently, the trained model parameters and number of data samples held by clients are sent back to the server, where the model parameters are aggregated. This process is repeated multiple times to train the model on the server.

For the prediction phase, the trained model parameters are distributed to client requiring predictions. The client applies the model to its locally held data to perform predictions.

\subsubsection{Parameter Integration Method}
This study employs FedAvg \cite{Federated_leaning_base} as the method for integrating model parameters. FedAvg performs weighted averaging of model parameters trained by each client, where the weights are determined by the number of data samples held by each client.

Specifically, let $\omega_k^t$ denote the model parameters trained by client $k$ at epoch $t$ and $n_k$ represent the number of data samples held by that client. The integrated model parameters $\omega^t$ after applying FedAvg are expressed as follows: 
\begin{equation} 
\omega^t = \sum_{k=1}^K \frac{n_k}{N} \omega_k^t \end{equation}
$N$ is the total number of data samples across all clients.

As shown in Figure \ref{fig:federated_learning_sequence}, the integrated model parameters $\omega^t$ from epoch $t$ are distributed to all clients at the beginning of epoch $t+1$. Each client then uses these parameters to train the model locally during the next epoch.

\subsubsection{Overview of Client-Side Learning}
\begin{figure}[htbp]
\begin{center}
\Description{The image of overview of client-side learning} 
\includegraphics[width=\linewidth]{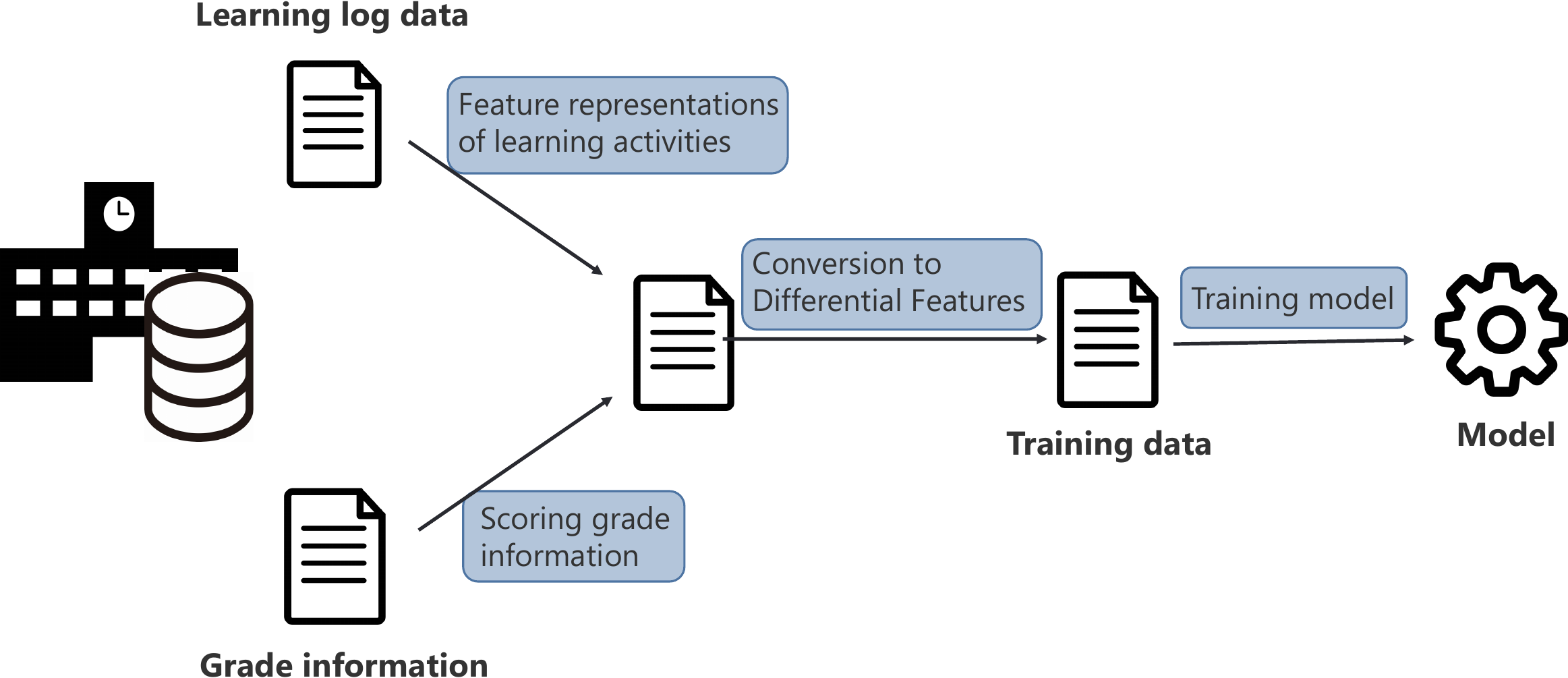} 
\caption{Overview of client-side learning} \label{fig:learning_outline} 
\end{center} 
\end{figure}

Figure \ref{fig:learning_outline} illustrates the part of the proposed method focused on client-side learning, which is extracted from the overall framework shown in Figure \ref{fig:big_picture}. Here, we briefly explain the client-side learning process.

First, learning log data are transformed into feature representations, while academic performance data are converted into scores. Subsequently, differential features are applied to these datasets to create relative data. Training on these relative data allows to obtain a generalized model.

The following sections provide a detailed explanation of each aspect.

\subsubsection{Learning Log Data in Digital Textbooks}
Digital textbook systems are educational platforms that allow learners to access and interact with learning materials through personal devices. While viewing the materials, learners can perform various actions, such as navigating forward or backward through pages, adding notes, or using markers. These interactions are recorded as learning log data and stored in a database.

\subsubsection{Feature Representation of Learning Activities}
\label{Feature Representation of Learning Activities}
\begin{figure}[htbp] 
\begin{center} 
\Description{The image of feature representation creation process using E2Vec} 
\includegraphics[width=\linewidth]{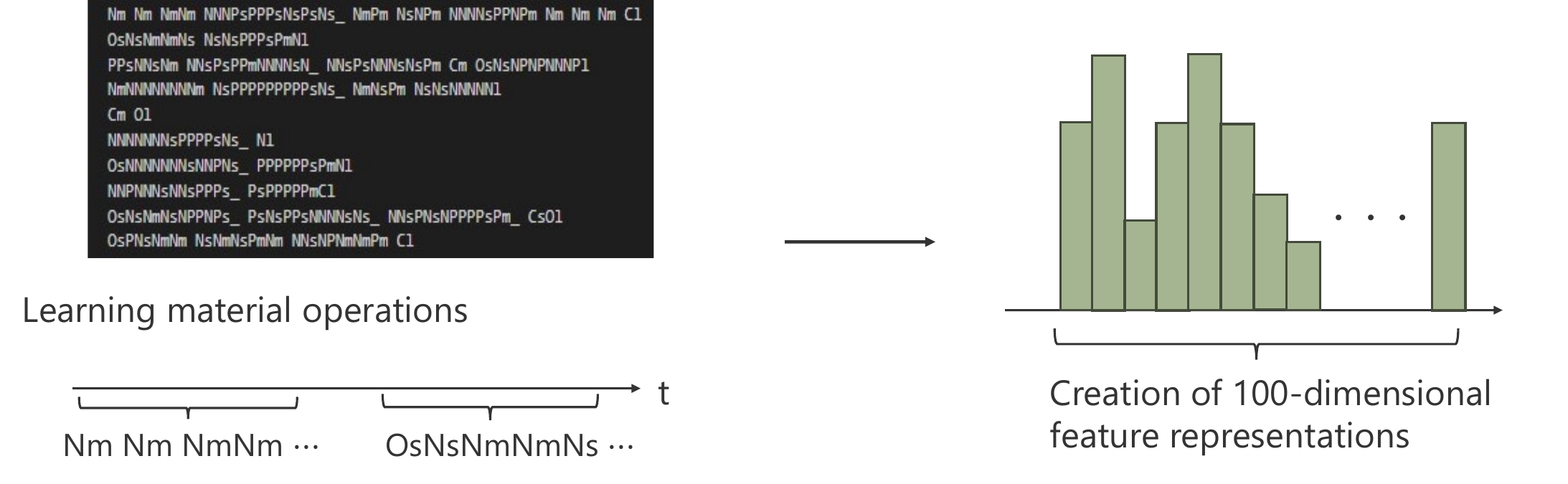} \caption{Feature representation creation using E2Vec} 
\label{fig:E2Vec} 
\end{center}
\end{figure}

Learning activities cannot be directly used to train ML models; thus, preprocessing the activities of each student into feature representations is necessary. To create feature representations, we adopted the distributed representation of learning material operations, E2Vec \cite{Miyazaki}. Figure \ref{fig:E2Vec} shows an overview of the feature representation creation process.

This method generates feature representations of learners by creating a sequence of symbols from learning logs while preserving the temporal order and time intervals between learning actions. E2Vec defines primitives, units, and actions corresponding to characters, words, and sentences in natural language processing. Learner’s log data are expressed as multiple actions with their distributed representations. These representations of actions are then aggregated using a method inspired by Bag of Visual Words \cite{Csurka2004Bagof}, resulting in 100-dimensional feature representations for each learner.

In the original EDM study by Miyazaki et al. \cite{Miyazaki}, the feature representations of each student were L2-normalized. However, in this study, normalization was omitted to account for the number of actions generated, enabling the features to be directly used for training.

\subsubsection{Scoring of Student Grades}
\label{sec:Scoring of grade information}
As described further in Section \ref{sec:Prediction_with_Regression_Model}, this study employs a regression model for at-risk students' prediction. Students' academic performance is represented by standard letter grades as points on a five-level scale: F, D, C, B, and A. To make these data compatible with the regression model, the grade points must be converted into numerical scores.

For each grade, let $x_1, x_2, x_3, x_4$, and $x_5$ denote the number of students with grades F, D, C, B, and A, respectively, and let $X$ represent the total number of students in the client. The converted grade value $G_m$ for grade $m$ is defined as:
\begin{equation} 
\label{equation:grade_scoring} 
G_m = \text{MaxScore} \times \frac{\sum_{j=1}^{m} x_j}{X} \ (m=1, 2, 3, 4, 5) 
\end{equation}

Additionally, considering the general criterion that students with total scores between 90\% and 100\% are assigned grade A, we adopt 0.95 as the value for MaxScore in this study.

\subsubsection{Ranking-Based Prediction Using Regression}
\label{sec:Prediction_with_Regression_Model}
This study employs a regression model to detect at-risk students. The feature representations are input into the regression model to obtain predicted academic performance values for each student. These prediction values are then sorted in ascending order to identify high-risk students in a ranking format, referred to as a \enquote{risk ranking} hereinafter.

\begin{figure}[htbp] 
\begin{center} 
\Description{The image of ranking creation method using a regression model}
\includegraphics[width=\linewidth]{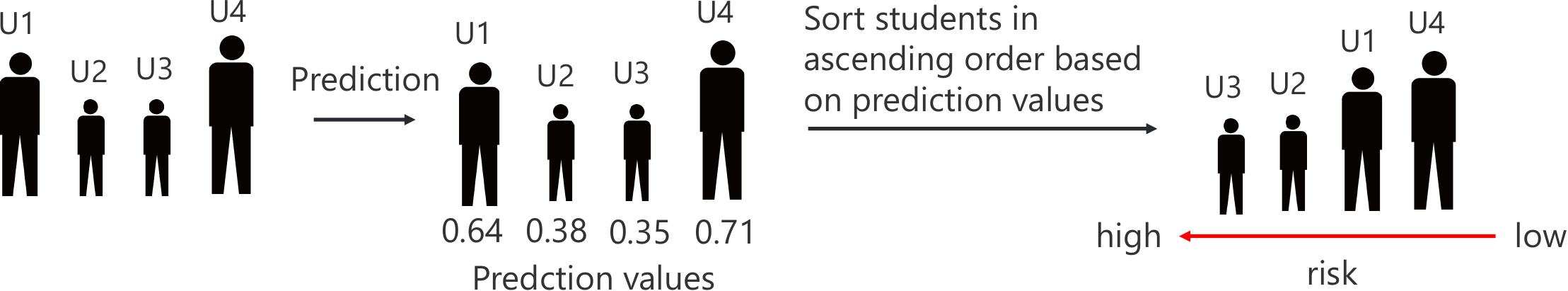} 
\end{center} 
\caption{Ranking creation method using a regression model} \label{fig:regression_model} 
\end{figure}

An example of the ranking creation using the regression model is shown in Figure \ref{fig:regression_model}. The prediction values in the figure are hypothetical and used for illustrative purposes.

The risk-ranking-based prediction using a regression model was adopted for the following three key reasons:
\begin{enumerate}
    \item \textbf{Compatibility with Differential Features}\\
    As described further in Section \ref{sec:Differential_Features}, the use of differential features enables to establish higher/lower relationships between students' grades. By employing a regression model, the extent of these differences can be explicitly learned, allowing the model to capture and utilize them.
    
    \item \textbf{Limitations of Classification Models}\\
    Classification models divide students into \textit{at-risk} and \textit{no-risk} groups. However, this complicates identifying no-risk students who are close to being at-risk or at-risk students who are closer to no-risk. This limitation has been observed in previous EDM studies~\cite{Svabensky2024detecting}. In contrast, this study utilizes a regression model to estimate predicted performance values. Ranking students based on these prediction values enables to better identify students near the boundary between the two groups.
    
    \item \textbf{Improved Generalization of the Predictions}\\
    The proposed method enables to create highly generalizable models for prediction. For example, consider a scenario where a course consists of 16 lecture sessions, and the goal is to detect at-risk students based on the data available until the 6th lecture for early prediction. If a model trained on data from all 16 lecture sessions is applied to early prediction data, classification models may struggle because of shifts in data distribution. Since classification relies on decision boundaries, the limited availability of data in early stages can lead to decreased confidence scores and increased misclassification rates. In contrast, a regression model applied in a ranking format mitigates this issue. While the absolute prediction values may fluctuate when trained on full-course data but applied to partial-course data, if this fluctuation occurs uniformly, the ranking order among students remains unaffected. This ensures that high-risk students can still be accurately identified.
    By leveraging a ranking-based approach with regression models, the proposed method eliminates the need to build separate models for early prediction, allowing models trained on complete data to be directly applied for detecting at-risk students at any stage of the course.
\end{enumerate}

\subsection{Differential Features}
\label{sec:Differential_Features}

This study employs \textit{differential features} to enhance the model's performance and generalizability. Such features are created by calculating the differences between feature representations and grade information of two students within the data held by a client. This approach generates data that represent the relative differences between students.

\begin{figure}[htbp]
\begin{center} 
\Description{The image of differential feature creation} 
\includegraphics[width=\linewidth]{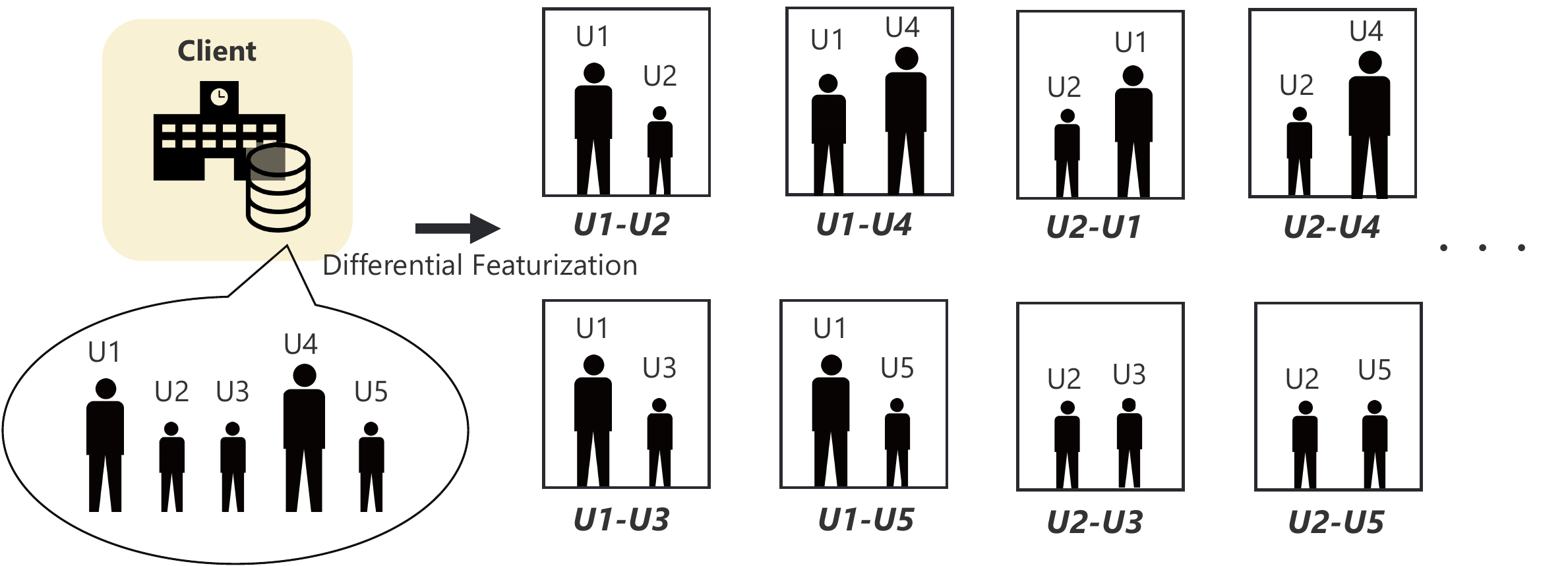}
\end{center}
\caption{Example of differential feature creation} \label{fig:differencial_features} 
\end{figure}

As an example, Figure \ref{fig:differencial_features} illustrates the application of differential features to a client that holds data for five students. The combinations shown in the figure represent only a subset of possible combinations for illustrative purposes and do not cover all potential pairings.

Specifically, let the set of students in a client be denoted~$\mathbf{S}$, the feature representation of student $i\ (i \in \mathbf{S})$ be $v_i$, and the grade of student $i$ after scoring based on Equation \ref{equation:grade_scoring} be $g_i$. Then, the feature representation $d_{ij}$ and grade information $e_{ij}$ after applying differential features are expressed as:
\begin{equation}
d_{ij} = v_i - v_j  \ (i \neq j, \  i,j \in \mathbf{S})
\end{equation}
\begin{equation}
e_{ij} = g_i - g_j  \ (i \neq j, \  i,j \in \mathbf{S})
\end{equation}

\subsubsection{Advantages of Differential Features}
Differential features have two main benefits:

\begin{enumerate}
    \item \textbf{Increase in Training Data}\\
    If a client holds data for $n$ students, the use of differential features expands the number of data points to $n(n-1)$. This expansion helps mitigate overfitting and bias when training the local model on clients with limited data because the increased data volume provides a richer dataset for training.
    
    \item \textbf{Improved Generalization by Utilizing Relative Values}\\
    Figure \ref{fig:differencial_merit} illustrates that introducing differential features enables the use of relative values. Without differential features, absolute feature values are used for training. This can bias the server model because of differences in feature distributions among clients. For instance, clients with different course structures (e.g., semester-based vs. quarter-based lectures) may exhibit substaintial differences in the features' absolute values due to varying interactions with learning materials.

    Instead, differential features use relative values, reducing the impact of such discrepancies and enabling the construction of a more generalized model. This approach improves the model's robustness and adaptability across clients with varying feature distributions.

    \begin{figure}[htbp] 
    \begin{center} 
    \Description{The image of utilization of relative values through differential features} 
    \includegraphics[width=\linewidth]{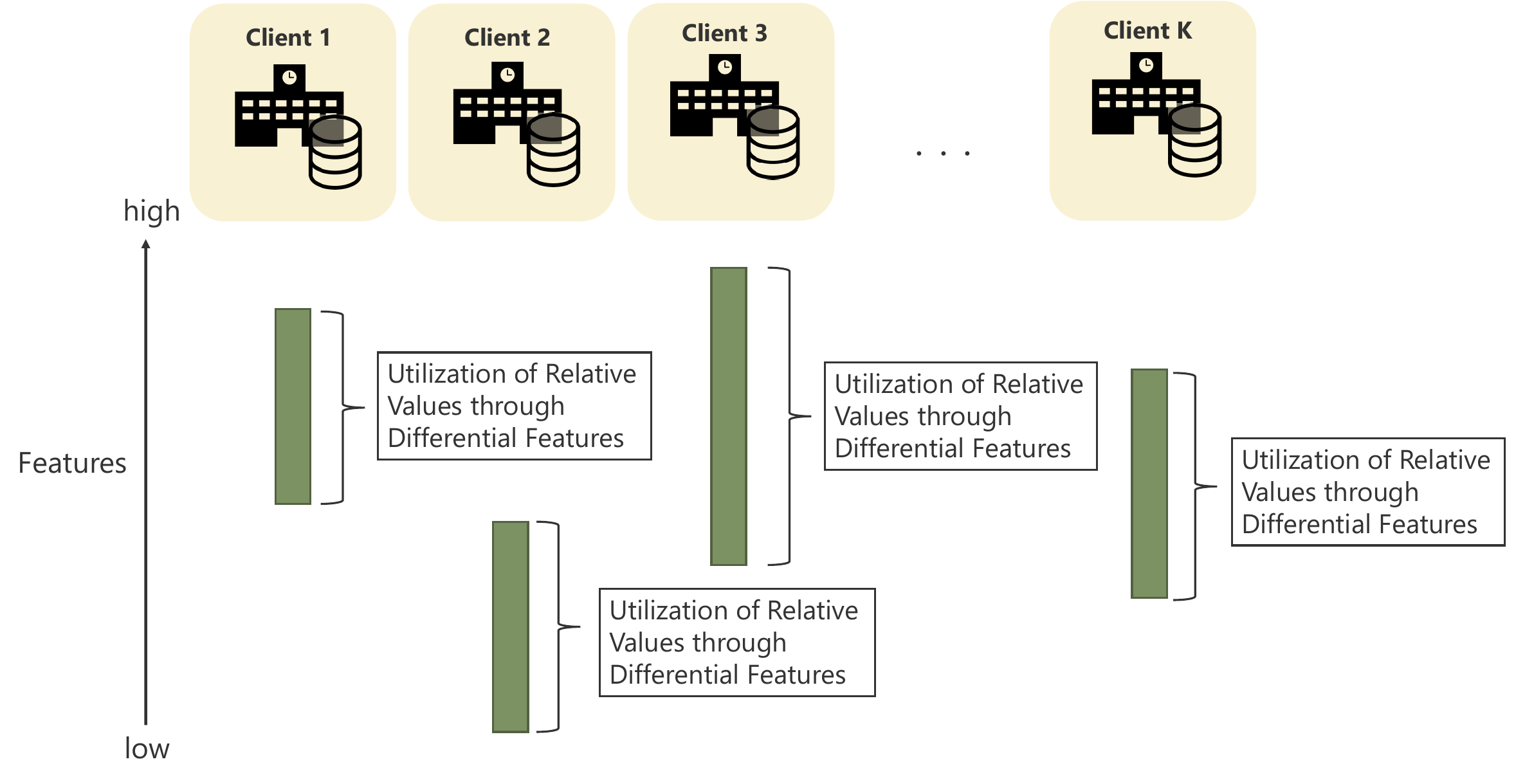} 
    \end{center} 
    \caption{Use of relative values in differential features (the graphs represent the distribution of feature values across different clients)}
    \label{fig:differencial_merit}
    \end{figure}
\end{enumerate}

\subsubsection{At-Risk Student Predction Using Differential Features}
As described in Section \ref{sec:Prediction_with_Regression_Model}, this study employs a regression model to rank students in terms of their risk level (in order from the highest to the lowest risk). However, with differential features, the regression model no longer outputs individual predictions for students but instead provides prediction values for the differences between two students (hereinafter, referred to as \enquote{pairwise difference scores}). Therefore, the method for obtaining individual prediction values when using differential features is described below.

Let $\mathbf{S}$ be the set of students in a client. Using the feature representation with differential features, $d_{ij}$, in the regression model yields a prediction value $p_{ij}$. However, since $p_{ij}$ represents the prediction value for the difference between two students, it cannot directly be used for at-risk detection. The individual prediction value $q_i$ for a student $i \ (i \in \mathbf{S})$ is derived as follows:
\begin{equation}
q_i = \sum_{j \in \mathbf{S}, j \neq i } p_{ij} \ ( i \in \mathbf{S} )
\end{equation}

\begin{figure}[htbp]
\begin{center} 
\Description{The image of example of deriving individual prediction values from differential feature based predictions} 
\includegraphics[width=\linewidth]{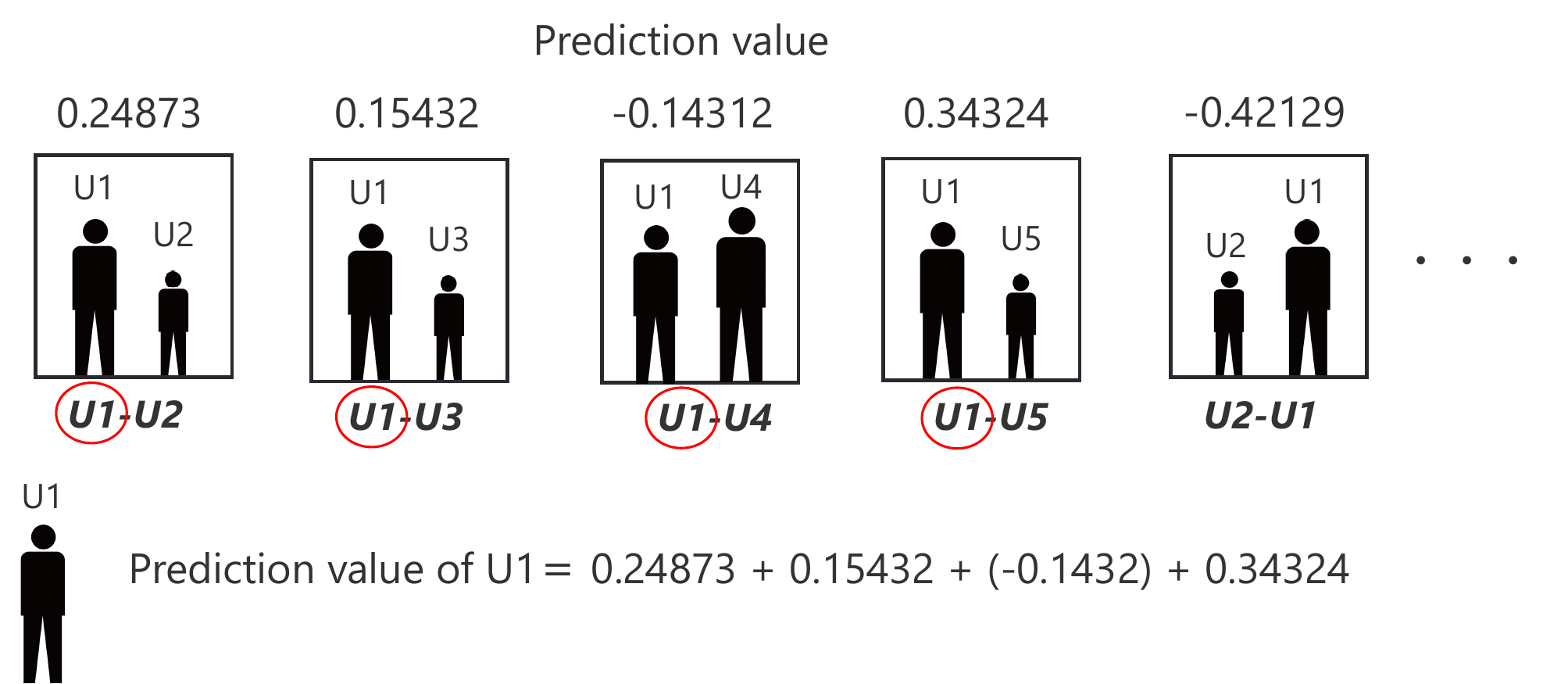}
\end{center} 
\caption{Example of deriving individual prediction values for a student U1 from differential-feature-based predictions} 
\label{fig:indivisual_value} 
\end{figure}

An example of calculating the individual prediction value for a specific student in a client with data for five students is illustrated in Figure \ref{fig:indivisual_value}. (The values are synthetic for illustrative purposes.)

After calculating the individual prediction values for all students, as described in Section \ref{sec:Prediction_with_Regression_Model}, the students are sorted in ascending order of their individual prediction values $q_i$. This creates a risk ranking, which can then be used to identify at-risk students based on their relative ranks.

It is important to note that the pairwise difference scores obtained from the regression model are used solely for internal model computation and are not provided to users (instructors and/or students, depending on the specific application). Instead, the output available to users is limited to a risk ranking of individual students. Therefore, users are not required to interpret the pairwise scores themselves, and the use of differential features does not hinder the interpretability of the system from the users' perspective.

\section{Experimental Evaluation}
\label{Experiments}
This section presents the comparison between the proposed method and baseline methods (defaults without the experimental condition). Additionally, it discusses the proposed method's early prediction capabilities and examines the corresponding results.

\subsection{Experimental Setup}
\subsubsection{Learning Log Data in the E-book Platform}
\begin{figure}[htbp]
\begin{center} 
\Description{The image of learning log data in BookRoll} 
\includegraphics[width=\linewidth]{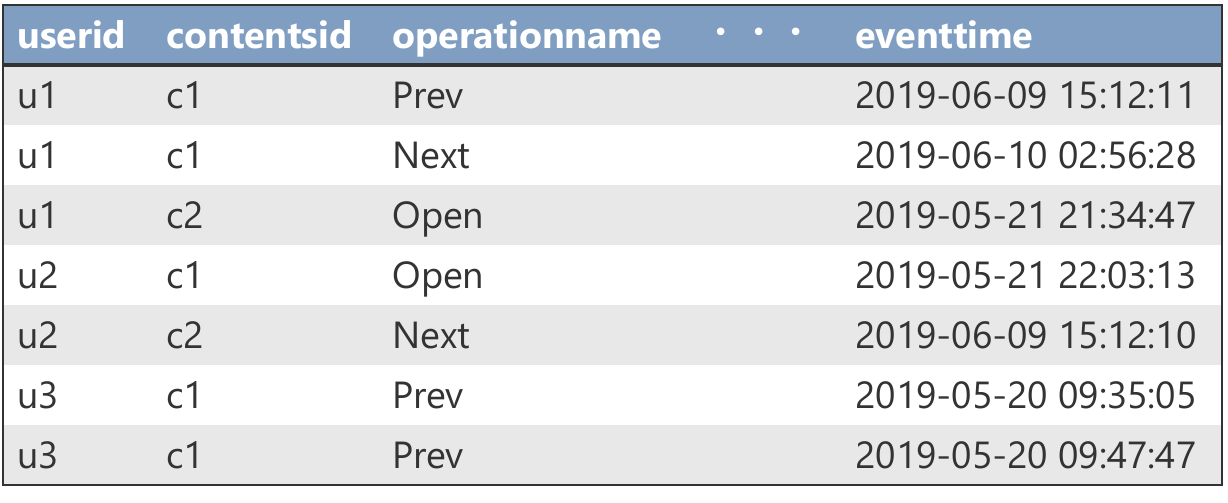}
\caption{Format of the learning logs in the BookRoll system} \label{fig:learning_log_data}
\end{center}
\end{figure}

BookRoll \cite{bookroll,BookRoll2} is a widely used system that allows students to access learning materials registered by instructors through their individual devices. When learners view the materials, control buttons are displayed alongside the content of the opened pages. These buttons have various functions, such as moving between pages and adding notes or markers. All these actions are recorded as log data and stored in a database. An example of the recorded log is shown in Figure \ref{fig:learning_log_data}.

The log data include the following information: IDs to identify the learner performing the operation and the material being accessed, the type of operation performed (denoted as \enquote{operation name}), and the timestamp of performing the operation (denoted as \enquote{event time}). Our study uses these learning log data for model training and prediction.

\subsubsection{Data and Clients in Federated Learning}
\label{sec:Data_and_Clients}
We collected and used a substantial dataset from \textit{four years} (eight semesters) of undergraduate courses at Kyushu University. A total of \textit{seven courses} were utilized for training and prediction, labeled from A to G. These courses span a diverse range of topics, learning formats, and academic terms, as detailed in Table~\ref{table:lectures_used_for_training_and_prediction}. 
In this study, data collection was conducted with the informed consent of the students. To ensure privacy protection, all collected data were anonymized and handled to prevent individual identification. Additionally, this study was approved by the institutional ethics committee.

\begin{table}[htbp]
\caption{Details of courses used for training and prediction}
\label{table:lectures_used_for_training_and_prediction}
\begin{center}
\resizebox{\linewidth}{!}{%
\begin{tabular}{c|l|l|l}
\textbf{Course} & \textbf{Topic} & \textbf{Format} & \textbf{Academic term}\\
\hline
A & Scheme & Lecture+Exercise & Quarter\\
B &  Security & Lecture  & Quarter\\
C &  Information and Communication & Lecture & Semester\\
D & Signal Processing & Lecture & Semester\\
E &  Programming & Exercise & Semester\\
F &  Artificial Intelligence Technology & Lecture & Semester\\
G &  Fortran & Lecture+Exercise & Semester\\
\end{tabular}%
}
\end{center}
\end{table}

The grade distribution of the training data used in this study is shown in Table \ref{table:learning_data}. This table presents the number of students receiving each grade (A, B, C, D, F) across different courses, along with the total number of students and the number of lecture weeks for each course. The training dataset includes \textit{1,136 students}, covering a wide range of courses and academic terms. In the \enquote{Course} column, the letter represents the course name, while the following number indicates the academic year in which the course was conducted.

Similarly, Table \ref{table:testing_data} shows the grade distribution for the hold-out test data used for prediction evaluation. In addition to student grades, this table includes the number of students classified as \enquote{At-risk} and \enquote{No-risk} based on their final grades. In this study, students are classified as at-risk if their grade is less than or equal to the grade of the student ranked 15th\footnote{The threshold of 15 was chosen to ensure that the number of at-risk students is sufficient for evaluating Top-n precision ($n$ = 15), see the next section.} from the bottom in their actual final grads, while the remaining students are classified as no-risk. This boundary is set arbitrarily for evaluation and is not used to differentiate between at-risk and no-risk during model training.

Finally, we note that the number of no-risk students (264) is more than twice that of at-risk students (127), resulting in a slightly imbalanced dataset. Nevertheless, this distribution is expected, since we assume to have more students who are not at risk.

In this study, the data in each row of Table \ref{table:learning_data} (except the summary row \enquote{Total}) were treated as a client, resulting in 12 clients for training. Subsequently, the trained model was applied to the 5 courses serving as hold-out test data (shown in Table \ref{table:testing_data}) to perform at-risk student prediction.

\begin{table}[tbp]
 \caption{Grade distribution and the number of lecture weeks in the \textit{training data}}
\label{table:learning_data}
\begin{center}
\resizebox{\linewidth}{!}{%
\begin{tabular}{l|rrrrr|r|r}
\textbf{Course} & \textbf{A} & \textbf{B} & \textbf{C} & \textbf{D} & \textbf{F} & \textbf{Students} & \textbf{Lectures}\\
\hline
A-2019 & 15 & 9 & 6 & 10 & 12 & 52 & 8\\
A-2020 & 22 & 23 & 5 & 3 & 7 & 60 & 7\\
A-2021 & 9 & 11 & 10 & 18 & 6 & 54 & 8\\
B-2019 & 30 & 103 & 28 & 1 & 1 & 163 & 8\\
C-2021-1 & 9 & 53 & 32 & 7 & 6 & 107 & 15\\
C-2021-2 & 15 & 88 & 37 & 26 & 9 & 175 & 15\\
D-2020 & 61 & 7 & 1 & 2 & 34 & 105 & 14\\
D-2021 & 60 & 3 & 6 & 4 & 33 & 106 & 15\\
E-2020-1 & 17 & 23 & 12 & 8 & 13 & 73 & 14\\
E-2020-2 & 0 & 2 & 8 & 21 & 25 & 56 & 15\\
F-2021 & 71 & 13 & 4 & 3 & 3 & 150 & 14\\
G-2021 & 26 & 3 & 3 & 0 & 3 & 35 & 16\\
\hline
Total & 335 & 338 & 152 & 103 & 152 & 1136 &\\
\end{tabular}%
}
\end{center}
\end{table}

\begin{table}[tbp]
\caption{Grade distribution and the number of lecture weeks in the separate hold-out \textit{test data} for prediction evaluation}
\label{table:testing_data}
\begin{center}
\resizebox{\linewidth}{!}{%
\begin{tabular}{l|rrrrr|cc|r}
\textbf{Course} & \textbf{A} & \textbf{B} & \textbf{C} & \textbf{D} & \textbf{F} & \textbf{No-risk} & \textbf{At-risk} & \textbf{Lectures}\\
\hline
A-2022 & 17 & 6 & 5 & \multicolumn{1}{|c}{22} & 2 & 28 & 24 & 8\\
\hline
B-2020 & 37 & 38 & \multicolumn{1}{|c}{12} & 2 & 4 & 75 & 18 & 7\\
C-2022-1 & 17 & 37 & \multicolumn{1}{|c}{34} & 4 & 4 & 54 & 42 & 15\\
\hline
D-2022 & 50 & 10 & 8 & 8 & \multicolumn{1}{|c|}{17} & 76 & 17 & 16\\
E-2021 & 3 & 16 & 8 & 4 & \multicolumn{1}{|c|}{26} & 31 & 26 & 16\\
\hline
Total & 124 & 107 & 67 & 40 & 53 & 264 & 127\\
\end{tabular}
}
\end{center}
\end{table}

\subsubsection{Evaluation Metrics}
We utilize a ranking-based approach where students are ordered in ascending order of their prediction values to create a risk ranking. To evaluate the propose method's performance, we use three ranking-specific evaluation metrics: Top-n precision, nDCG, and PR-AUC:
\begin{enumerate}
    \item \textbf{Top-n Precision}\\
    Top-n precision indicates the proportion of students who are actually at-risk among the top-n students predicted to incur the highest risk. In this study, we evaluate Top-n precision with four different settings: $n =$ \{5, 10, 15, At-risk\}. Here, the \enquote{At-risk} refers to the actual number of students classified as at-risk in the test data.
    \item \textbf{Normalized Discounted Cumulated Gain (nDCG)}\\  
    As described in the reference paper \cite{ndcg}, nDCG is a ranking-specific metric used to compare the predicted ranking order based on prediction values with the actual ranking order based on the students’ grades. To compute nDCG, each student must be assigned a value. Since higher-risk students are ranked higher in this study, the values assigned to students must increase as their risk level increases. Therefore, we utilize the score \(G_m\) derived from Equation \ref{equation:grade_scoring} and assign each student a value of \(1 - G_m\ (m = 1, 2, 3, 4, 5)\) to calculate nDCG.
    \item \textbf{Area Under the Precision-Recall Curve (PR-AUC)}\\  
    The Precision-Recall (PR) curve is commonly used in EDM research \cite{PR-AUC1,PR-AUC2} to evaluate predictive model performance. It plots Top-n precision (vertical axis) against Top-n recall (horizontal axis) as $n$ varies, illustrating their relationship. The area under this curve (i.e., PR-AUC) quantifies model performance, with higher values indicating better balance between precision and recall.
\end{enumerate}

For a robust evaluation, a single evaluation result from one model would be insufficient. Therefore, we conducted training 10 times and calculated the average of the evaluation metrics obtained from the models. This approach provides a more reliable assessment of the model's performance.

\subsubsection{Regression Model}
A neural network was employed as the regression model during training. This neural network consists of two hidden layers: the first layer comprises 50 nodes and the second layer 10 nodes. To prevent overfitting, we applied dropout between the hidden layers, with a dropout rate of 20\%. The activation function in each hidden layer is ReLU, commonly used in similar EDM contexts \cite{generalizability_3, using_relu1}.

\subsubsection{Comparisons of Experimental and Baseline Conditions}
\label{sec:Comparison_Methods}
We conducted two types of comparative experiments: 
\begin{enumerate}
    \item \textbf{Performance comparison between \enquote{Federated Learning} and \enquote{Centralized Machine Learning}}\\
    Centralized ML is the baseline method in which the data from all clients are collected in a single location, with potential privacy concerns. The model is trained using the aggregated data.
    
    \item \textbf{Performance comparison between \enquote{With Differential Features} and \enquote{Without Differential Features}}\\
    Without differential features, applying the regression model to the feature representations yields individual prediction values for each student. As described in Section \ref{sec:Prediction_with_Regression_Model}, students can be easily ranked in order of risk by sorting them in ascending order of their individual prediction values.  
\end{enumerate}

\subsection{Experimental Results}
\subsubsection{Federated vs. Centralized Learning}
\label{Federated_vs_Centralized_Learning}

\begin{table*}[tb]
\caption{Proposed Method (using Federated Learning) vs. Baseline Method 1 (using Centralized ML)}
\label{table:Federated_or_not}
\begin{center}
\begin{tabular}{c|c|cccc|c|c}
\multirow{2}{*}{\textbf{Test Data}} & \multirow{2}{*}{\textbf{Method}} & \multicolumn{4}{c|}{\textbf{Top-n precision}} & \multirow{2}{*}{\textbf{nDCG}} & \multirow{2}{*}{\textbf{PR-AUC}} \\
 & & $n=5$ & $n=10$ & $n=15$ & $n=\text{At-risk}$ & & \\
\hline
\multirow{2}{*}{A-2022}&\textbf{Proposed Method}& 0.96 & 0.83 & 0.80 & \textbf{0.63} & 0.83 & \textbf{0.75} \\
&Baseline Method 1& \textbf{1.00} & \textbf{0.85} & \textbf{0.81} & \textbf{0.63} & \textbf{0.84} & \textbf{0.75} \\

\hline
\multirow{2}{*}{B-2020}&\textbf{Proposed Method}& \textbf{0.72} & \textbf{0.56} & \textbf{0.43} & \textbf{0.41} & \textbf{0.72} & \textbf{0.46} \\
&Baseline Method 1& 0.64 & 0.53 & 0.41 & 0.37 & 0.69 & 0.43 \\

\hline
\multirow{2}{*}{C-2022-1}&\textbf{Proposed Method}& \textbf{1.00} & 0.93 & \textbf{0.79} & \textbf{0.72} & \textbf{0.87} & \textbf{0.78} \\
&Baseline Method 1& \textbf{1.00} & \textbf{0.95} & \textbf{0.79} & 0.70 & 0.85 & \textbf{0.78} \\

\hline
\multirow{2}{*}{D-2022}&\textbf{Proposed Method}& \textbf{0.80} & \textbf{0.90} & \textbf{0.84} & \textbf{0.79} & \textbf{0.95} & \textbf{0.83} \\
&Baseline Method 1& \textbf{0.80} & \textbf{0.90} & 0.83 & \textbf{0.79} & \textbf{0.95} & \textbf{0.83} \\

\hline
\multirow{2}{*}{E-2021}&\textbf{Proposed Method}& \textbf{1.00} & \textbf{0.97} & \textbf{0.83} & \textbf{0.64} & \textbf{0.86} & \textbf{0.78} \\
&Baseline Method 1& 0.98 & 0.96 & \textbf{0.83} & 0.63 & 0.85 & 0.77 \\
\end{tabular}
\end{center}
\end{table*}

\begin{table*}[tb]
\caption{Proposed Method (With Differential Features) vs. Baseline Method 2 (Without Differential Features)}
\label{table:difference_or_not}
\begin{center}
\begin{tabular}{c|c|cccc|c|c}
\multirow{2}{*}{\textbf{Test Data}} & \multirow{2}{*}{\textbf{Method}} & \multicolumn{4}{c|}{\textbf{Top-n precision}} & \multirow{2}{*}{\textbf{nDCG}} & \multirow{2}{*}{\textbf{PR-AUC}} \\
 & & $n=5$ & $n=10$ & $n=15$ & $n=\text{At-risk}$ & & \\
\hline
\multirow{2}{*}{A-2022}&\textbf{Proposed Method}& \textbf{0.96} & \textbf{0.83} & \textbf{0.80} & \textbf{0.63} & \textbf{0.83} & \textbf{0.75} \\
&Baseline Method 2& 0.88 & 0.83 & 0.73 & 0.61 & 0.80 & 0.70 \\

\hline
\multirow{2}{*}{B-2020}&\textbf{Proposed Method}& \textbf{0.72} & \textbf{0.56} & \textbf{0.43} & \textbf{0.41} & \textbf{0.72} & \textbf{0.46} \\
&Baseline Method 2& 0.58 & 0.47 & 0.35 & 0.31 & 0.62 & 0.37 \\

\hline
\multirow{2}{*}{C-2022-1}&\textbf{Proposed Method}& \textbf{1.00} & \textbf{0.93} & 0.79 & \textbf{0.72} & \textbf{0.87} & \textbf{0.78} \\
&Baseline Method 2& 0.96 & 0.92 & \textbf{0.85} & 0.65 & 0.82 & 0.75 \\

\hline
\multirow{2}{*}{D-2022}&\textbf{Proposed Method}& 0.80 & \textbf{0.90} & \textbf{0.84} & \textbf{0.79} & \textbf{0.95} & \textbf{0.83} \\
&Baseline Method 2& \textbf{0.86} & 0.85 & 0.79 & 0.76 & 0.93 & 0.82 \\

\hline
\multirow{2}{*}{E-2021}&\textbf{Proposed Method}& \textbf{1.00} & \textbf{0.97} & \textbf{0.83} & \textbf{0.64} & \textbf{0.86} & \textbf{0.78} \\
&Baseline Method 2& 0.92 & 0.80 & 0.71 & 0.60 & 0.81 & 0.68 \\
\end{tabular}
\end{center}
\end{table*}

First, we compared \enquote{Federated Learning} and \enquote{Centralized Machine Learning}. To ensure stable and uniform conditions for comparison, all evaluations used differential features. The comparison results are shown in Table \ref{table:Federated_or_not}.

These results demonstrate that the Proposed Method (using federated learning) achieves almost the same performance as Baseline Method 1 (using centralized machine learning) across all test data.
Specifically, in terms of nDCG, the performance degradation of the Proposed Method is at most 0.01 in A-2022, while maintaining comparable or superior performance in other test data. Furthermore, for PR-AUC and Top-n precision ($n=$ At-risk), the Proposed Method consistently demonstrates performance that is equal to or better than Baseline Method 1 across all test data.

Therefore, federated learning enables at-risk student prediction with performance comparable to that of centralized machine learning, with the added benefit of more strongly preserved privacy.

\subsubsection{With vs. Without Differential Features}
Next, we compared \enquote{With Differential Features} and \enquote{Without Differential Features}. To ensure stable and uniform conditions for comparison, all evaluations used federated learning. The comparison results are presented in Table \ref{table:difference_or_not}.

The Proposed Method (with differential features) consistently outperformed Baseline Method 2 (without differential features) across all test data in terms of both nDCG and PR-AUC metrics. This indicates that introducing differential features enables more accurate detection of at-risk students, at least in the context of e-book log data.

This improvement is primarily due to two key advantages of differential features: (1) data augmentation, which increases the amount of training data per client and stabilizes learning, reducing overfitting; and (2) relative value utilization, which mitigates differences in feature distributions across clients, improving generalization. These benefits allow the model to maintain stable performance even in heterogeneous educational environments. However, while differential features substantially contribute to these improvements, other potential factors might have also influenced the results.

The findings suggest that introducing differential features enhances not only the performance but also the generalizability of the prediction model, enabling it to achieve stable performance across various test data.

\subsubsection{Application to Early Prediction}
\begin{figure*}[htbp]
\Description{The image of relationship between lecture sessions and PR-AUC in several courses} 
  \centering
  \begin{minipage}{0.32\linewidth}
    \centering
    \Description{The image of relationship between lecture sessions and PR-AUC in A-2022} 
    \includegraphics[width=\linewidth]{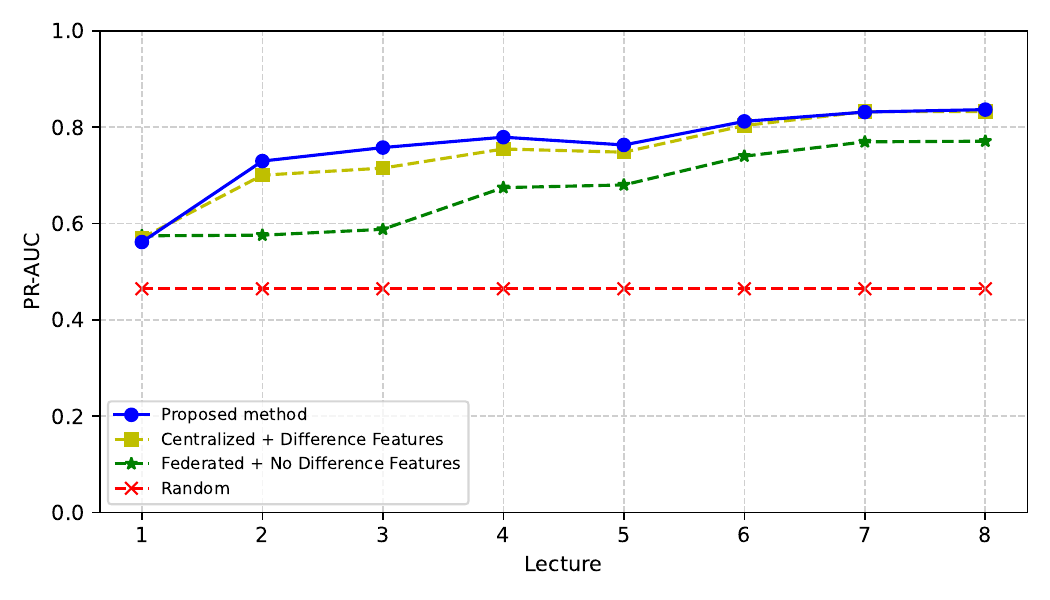}
    \caption{Relationship between lecture sessions and PR-AUC in course A-2022}
    \label{fig:Relationship in Lecture A-2022}
  \end{minipage}%
  \hfill
  \begin{minipage}{0.32\linewidth}
    \centering
    \Description{The image of relationship between lecture sessions and PR-AUC in B-2020} 
    \includegraphics[width=\linewidth]{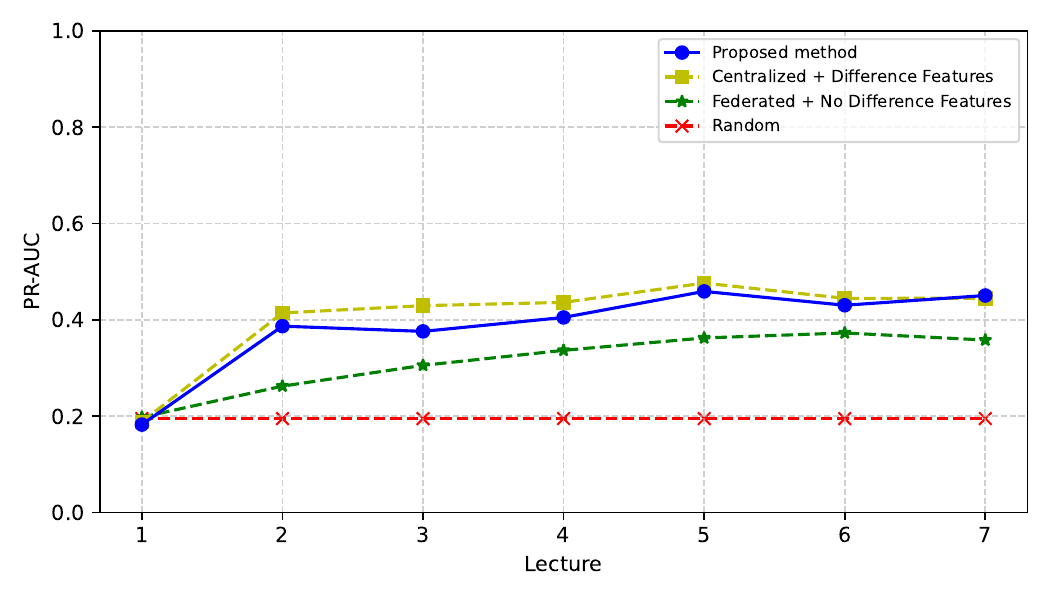}
    \caption{Relationship between lecture sessions and PR-AUC in course B-2020}
    \label{fig:Relationship in Lecture B-2020}
  \end{minipage}%
  \hfill
  \begin{minipage}{0.32\linewidth}
    \centering
    \Description{The image of relationship between lecture sessions and PR-AUC in C-2022-1} 
    \includegraphics[width=\linewidth]{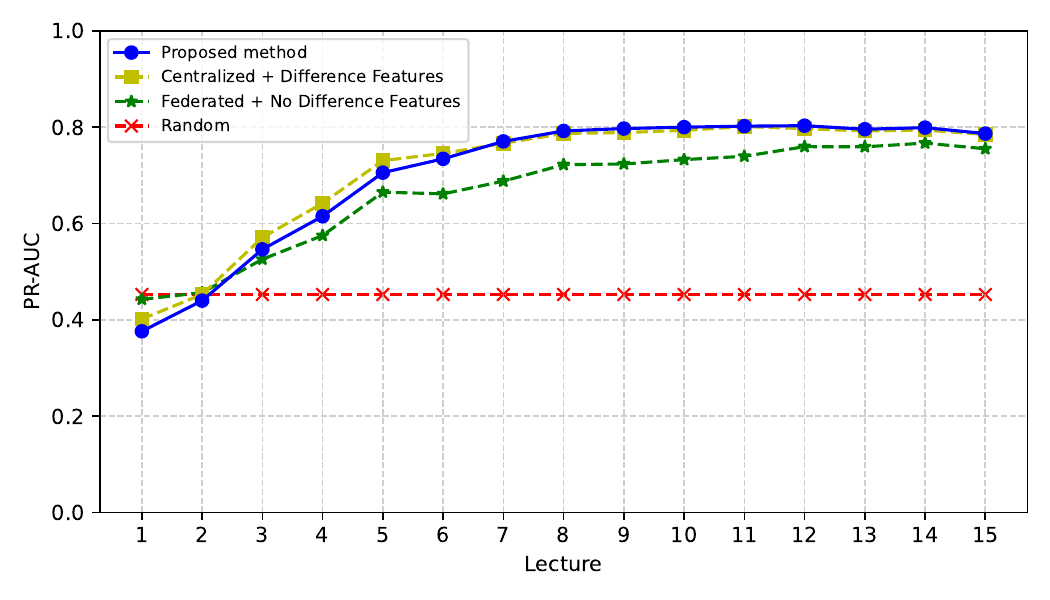}
    \caption{Relationship between lecture sessions and PR-AUC in course C-2022-1}
    \label{fig:Relationship in Lecture C-2022-1}
  \end{minipage}\\[5mm]
  
  \begin{minipage}{0.32\linewidth}
    \centering
    \Description{The image of relationship between lecture sessions and PR-AUC in D-2022} 
    \includegraphics[width=\linewidth]{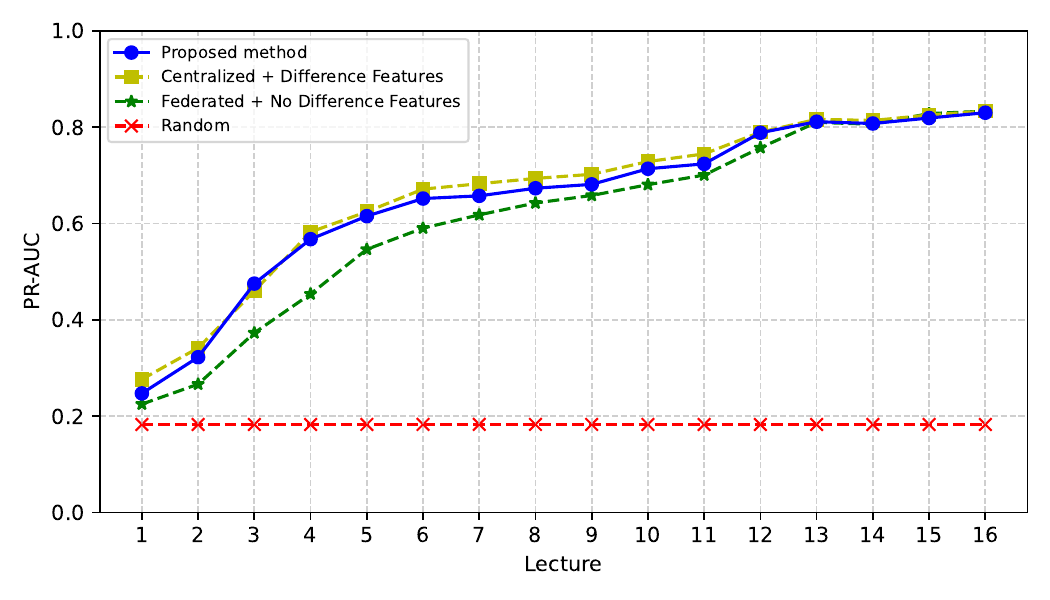}
    \caption{Relationship between lecture sessions and PR-AUC in course D-2022}
    \label{fig:Relationship in Lecture D-2022}
  \end{minipage}%
  \hspace{3mm}
  \begin{minipage}{0.32\linewidth}
    \centering
    \Description{The image of relationship between lecture sessions and PR-AUC in E-2021} 
    \includegraphics[width=\linewidth]{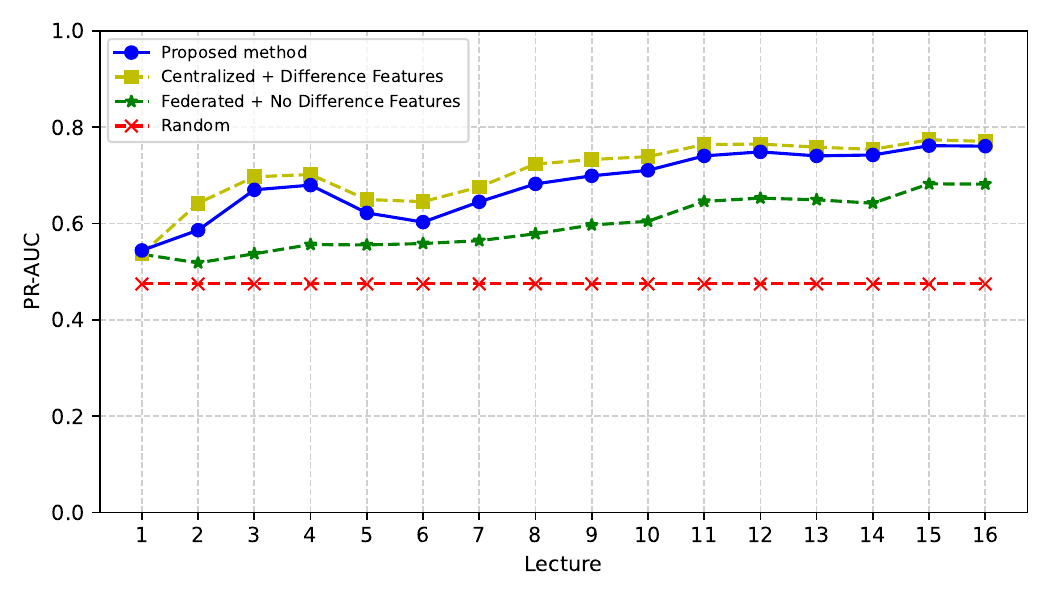}
    \caption{Relationship between lecture sessions and PR-AUC in course E-2021}
    \label{fig:Relationship in Lecture E-2021}
  \end{minipage}
\end{figure*}

In the previous sections, predictions were conducted using data collected after the completion of all lecture sessions. However, in practice, the early prediction of at-risk students is crucial \cite{sukrit_early_prediction}. 
Therefore, we investigated the prediction performance when using learning logs obtained from lecture sessions up to a certain point. The results are shown in Figures \ref{fig:Relationship in Lecture A-2022} to \ref{fig:Relationship in Lecture E-2021}.

\begin{table*}[tb]
\caption{Evaluation of Risk Ranking in Early Prediction}
\label{table:early}
\begin{center}
\begin{tabular}{c|c|cccc|c|c}
\multirow{2}{*}{\textbf{Test Data}} & \multirow{2}{*}{\textbf{Method}} & \multicolumn{4}{c|}{\textbf{Top-n precision}} & \multirow{2}{*}{\textbf{nDCG}} & \multirow{2}{*}{\textbf{PR-AUC}} \\
 & & $n=5$ & $n=10$ & $n=15$ & $n=\text{At-risk}$ & & \\
 
\hline
A-2022 & \multirow{5}{*}{\textbf{Proposed Method}} & 1.00 & 0.82 & 0.77 & 0.71 & 0.81 & 0.78\\
B-2020 &  & 0.64 & 0.48 & 0.41 & 0.37 & 0.68 & 0.40 \\
C-2022-1 & & 1.00 & 0.97 & 0.92 & 0.73 & 0.84 & 0.79 \\
D-2022 &  & 0.78 & 0.82 & 0.73 & 0.69 & 0.87 & 0.67 \\
E-2021 &  & 0.90 & 0.80 & 0.71 & 0.58 & 0.80 & 0.68 \\
\end{tabular}
\end{center}
\end{table*}

In these figures, the horizontal axis represents the number of completed lecture sessions, while the vertical axis indicates the PR-AUC evaluation score. The figures illustrate how the prediction performance improves as the number of lecture sessions used in the prediction increases. The blue points represent the results obtained using the proposed method; the yellow points represent the results of Baseline Method 1; the green points represent the results of Baseline Method 2, and the red points represent the results obtained by arranging the students in a random order.

The results show that for the quarter-based lectures, A-2022 and B-2020, predictions at the end of the 4th lecture achieved performance almost equivalent to that of predictions made after the final lecture. For semester-based lectures, C-2022-1 achieved comparable performance to the final prediction at the end of the 8th lecture, while D-2022 showed continuous improvement in prediction performance as lecture sessions used for prediction increased. For E-2021, while performance temporarily declined, it improved as lecture sessions used in the prediction increased.

In summary, for A-2022, B-2020, and C-2022-1, detection performance after half of the total lectures was comparable to performance after all lectures. Therefore, for early prediction, the evaluation was conducted using data up to the 4th lecture for A-2022 and B-2020 and up to the 8th lecture for C-2022-1, D-2022, and E-2021. The results are summarized in Table~\ref{table:early}. These results indicate that high performance in at-risk prediction can still be achieved in early prediction scenarios.

For all test data except B-2020, Top-n precision ($n=15$) exceeded 0.7. Thus, when extracting the top 15 high-risk students from the risk ranking, at least 10 of them were accurately classified as at-risk in their final grades. These findings demonstrate the effectiveness of the proposed method for early prediction and its ability to accurately identify high-risk students early in the lecture series.

\subsubsection{Risk Rankings in Early Prediction}
\begin{figure*}[tb]
\Description{The image of relationship between grades and rankings in early prediction for 5 courses} 
  \centering
  \begin{minipage}{0.32\linewidth}
    \centering
    \Description{The image of relationship between grades and rankings in early prediction for A-2022} 
    \includegraphics[width=\linewidth]{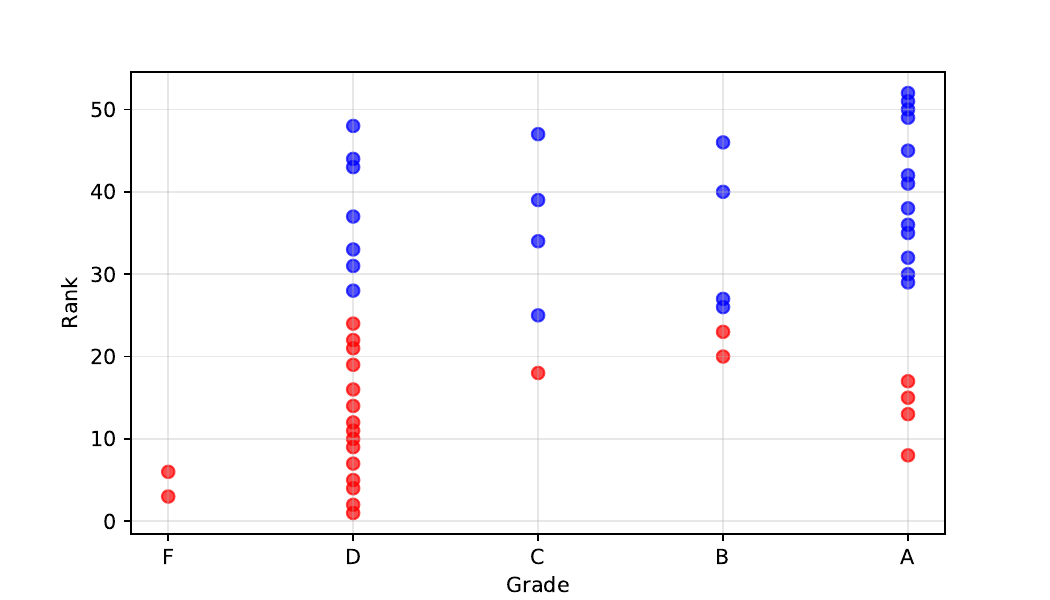}
    \caption{Relationship of grades and rankings in early prediction for A-2022}
    \label{fig:early_A-2022_At-risk Rank}
  \end{minipage}%
  \hfill
  \begin{minipage}{0.32\linewidth}
    \centering
    \Description{The image of relationship between grades and rankings in early prediction for B-2020} 
    \includegraphics[width=\linewidth]{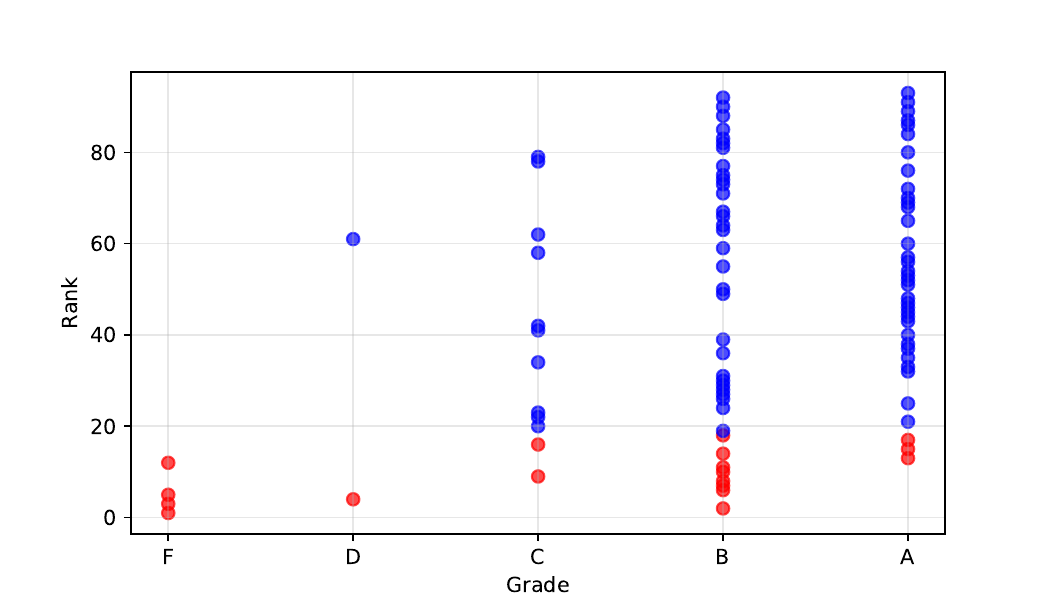}
    \caption{Relationship of grades and rankings in early prediction for B-2020}
    \label{fig:early_B-2020_At-risk Rank}
  \end{minipage}%
  \hfill
  \begin{minipage}{0.32\linewidth}
    \centering
    \Description{The image of relationship between grades and rankings in early prediction for C-2022-1} 
    \includegraphics[width=\linewidth]{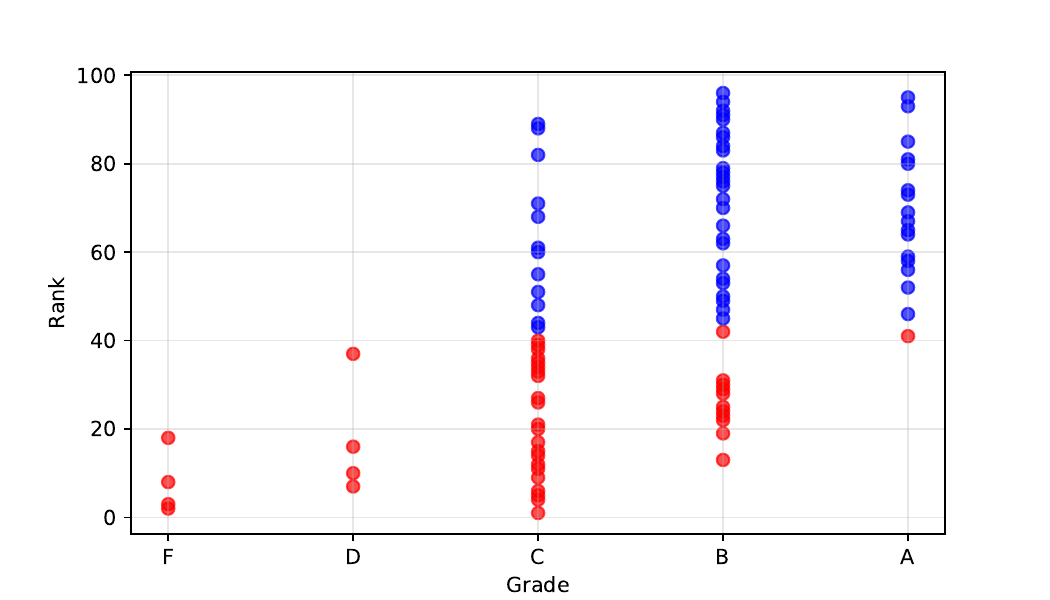}
    \caption{Relationship of grades and rankings in early prediction for C-2022-1}
    \label{fig:early_C-2022-1_At-risk Rank}
  \end{minipage}\\[5mm]

  \begin{minipage}{0.32\linewidth}
    \centering
    \Description{The image of relationship between grades and rankings in early prediction for D-2022} 
    \includegraphics[width=\linewidth]{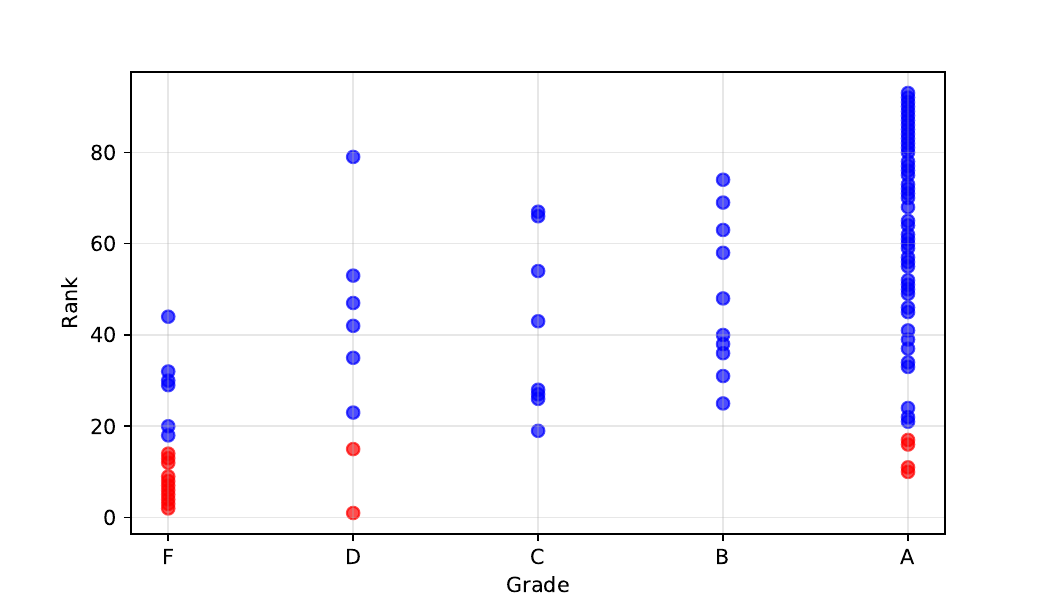}
    \caption{Relationship of grades and rankings in early prediction for D-2022}
    \label{fig:early_D-2022_At-risk Rank}
  \end{minipage}%
  \hspace{3mm}
  \begin{minipage}{0.32\linewidth}
    \centering
    \Description{The image of relationship between grades and rankings in early prediction for E-2021} 
    \includegraphics[width=\linewidth]{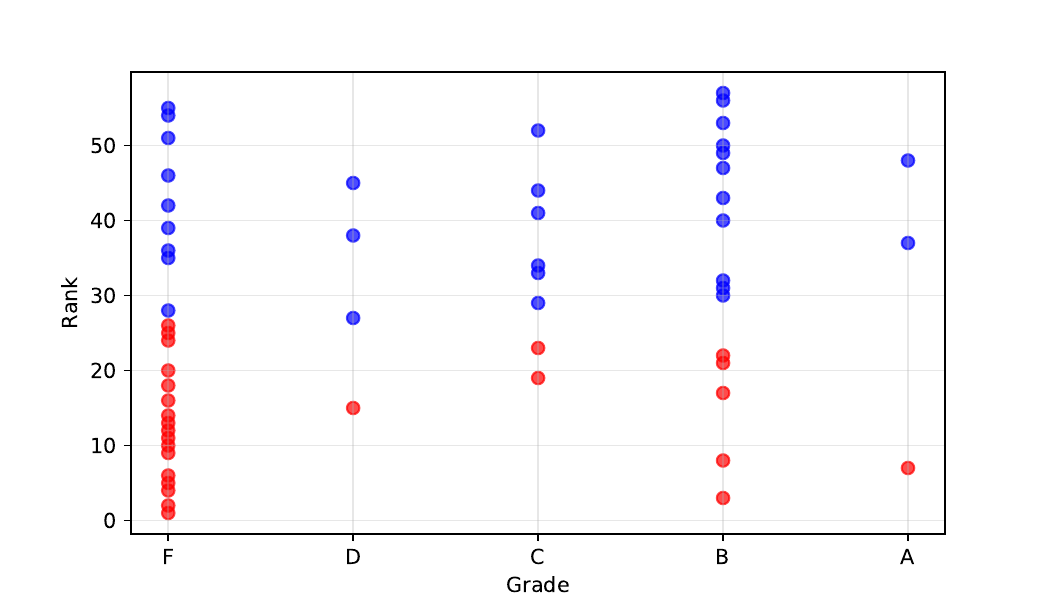}
    \caption{Relationship of grades and rankings in early prediction for E-2021}
    \label{fig:early_E-2021_At-risk Rank}
  \end{minipage}
\end{figure*}

The visualized risk rankings in early prediction using the proposed method are shown in Figures \ref{fig:early_A-2022_At-risk Rank} to \ref{fig:early_E-2021_At-risk Rank}.

In these figures, the horizontal axis represents the students' grades (F, D, C, B, A), while the vertical axis represents their ranks in the risk ranking. The dots are color-coded: students identified as at-risk (corresponding to the number of at-risk students in Section \ref{sec:Data_and_Clients}) are marked red, while all other students are marked blue. This color coding enables an intuitive understanding of the relationship between students’ grades and their ranks in the risk ranking.

In most test data, students with lower grades tend to be represented by red dots, which indicates a higher proportion of at-risk students among those with poor academic performance.
These results suggest that the proposed method effectively captures the relationship between academic performance and risk, consistently identifying students with lower grades as high-risk. This demonstrates that the generated risk rankings align well with the expected academic trends, reinforcing the validity of the model's predictions.

\subsection{Discussion and Limitations}
\label{sec:limitations}
\subsubsection{Data Source and Generalizability}
\label{sec:Data Source and Generalizability}
Although the proposed approach has been evaluated with learning logs only from BookRoll, our method is applicable in other learning management systems. We publicly provide our modeling code (see Section \ref{Conclusion_and_Future_Work}) for others to adopt or extend to other types of learning logs. As a result, further research may adapt our methods to other systems and datasets.
However, since all data in this study originates from a single institution and platform, the generalizability of the proposed method to different institutional contexts remains an open question. 

\subsubsection{Communication and Computational Constraints}
Regarding federated learning, this study did not consider the network communication overhead or the client devices’ computational constraints. However, federated learning inherently introduces challenges related to network communication costs and device capabilities. These limitations must be considered for real-world deployment, and future work should explore lightweight models and communication-efficient strategies.

\subsubsection{Model Interpretability}
\label{sec:Model Interpretability}
While our results show that federated learning achieves prediction performance comparable to that of centralized learning (Section \ref{Federated_vs_Centralized_Learning}), the interpretability of the resulting model has not been fully explored in this study. In particular, it remains unclear which specific input features contribute the most to the risk predictions. Investigating the influence of individual features within the differential features will be an important direction for future work, especially for improving the model’s transparency in practical educational settings.

\newpage

\subsubsection{Robustness to Non-IID Data}
An important factor to consider in federated learning is whether the model can perform well under non-independent and identically distributed (non-IID) data.
This study introduced differential features to address discrepancies and skew in feature distributions across clients—one aspect of non-IID data in federated learning.
Notably, differential features transform both the input features and student outcome labels (i.e., grade information) into pairwise differences between students. While our method was not originally designed to address label distribution skew—another important aspect of non-IID data—it may incidentally help mitigate this issue. Although we did not evaluate this effect, we recognize it as a potential secondary benefit of using differential features and consider it an important direction for future work.

\subsubsection{Assumptions Behind Differential Features}
While our results show improved performance with differential features, we implicitly assume that these features mitigate discrepancies in feature distributions across clients. However, this assumption has not yet been explicitly validated for a general setting, and further investigation in other educational contexts is needed to confirm whether the observed improvements are indeed attributable to reduced inter-client variability in feature distributions.

\subsection{Implications for Educational Practice}
Our approach offers practical value for both instructors and students. From the instructor’s perspective, the risk rankings can inform individualized support strategies, such as prioritizing interventions for high-risk students or organizing differentiated instruction. From the students’ perspective, understanding their relative risk status may help them reflect on their learning behavior and take proactive steps toward improvement. 

\section{Conclusion and Future Work}
\label{Conclusion_and_Future_Work}

This study proposed a method that (1) applies federated learning in EDM to enable privacy-preserving prediction modeling and (2) leverages differential features to use relative values between clients, resulting in a generalizable and high-performing model. Additionally, to effectively utilize differential features, we proposed a method that scores grades, employs regression, and calculates individual prediction values from pairwise difference scores to generate risk rankings.

The evaluation of the proposed method demonstrated the following novel contributions:
\begin{itemize} 
    \item Federated learning achieves at-risk student prediction performance comparable to that of centralized ML, while benefiting from increased privacy protection.
    \item Introducing differential features improves the performance of at-risk student prediction compared to the baseline without differential features.
    \item Even when using data from only half of the lecture sessions, the proposed method achieves high performance in at-risk student prediction, which demonstrates its applicability to early prediction scenarios.
\end{itemize}

\subsection{Open Research Challenges}

As a result of this novel study, there are several open research challenges in exploring federated learning and differential features in EDM. Specifically, the following topics are identified as areas for future work:
\begin{enumerate} 
    \item \textbf{Developing a More Generalizable Model}\\  
    As discussed in Section \ref{sec:Data Source and Generalizability}, the proposed approach is currently designed for learning log data from BookRoll. To develop a more generalizable and robust model, further work could investigate its applicability to different types of learning logs and educational datasets, exploring methods for adapting the model to diverse data sources and system architectures.
    
    \item \textbf{Alternative Integration Methods in Federated Learning}\\  
    This study employed the commonly used method called FedAvg for federated parameter integration. While FedAvg is a widely adopted baseline, other integration methods such as FedOpt \cite{FedOpt} and SCAFFOLD \cite{SCAFFOLD} may offer advantages in terms of convergence speed, stability, or model performance. Future work could explore the impact of these alternative methods on the effectiveness of the proposed approach.

    \item \textbf{Identifying Actions Influencing Risk Levels}\\ 
    Since the risk rankings are generated based on the learning logs, further investigation is required to identify which specific student actions contribute to lower or higher risk, as discussed in Section \ref{sec:Model Interpretability}.
 
\end{enumerate}

\subsection{Availability of Research Code}

To support the replicability of the findings—an aspect valued by the EDM community~\cite{Haim2023edm}—we have made the code used to produce the results in this paper publicly available at:
\begin{center}
\url{https://github.com/limu-research/2025-EDM-FL}.
\end{center}

\section{Acknowledgments}
This work was supported by JST CREST Grant Number JPMJCR22D1 and JSPS KAKENHI Grant Number JP22H00551, Japan.

\bibliographystyle{abbrv}
\bibliography{ref}


\balancecolumns
\end{document}